\documentclass[conference]{IEEEtran}
\usepackage{times}

\usepackage[numbers]{natbib}
\usepackage{multicol}
\usepackage[bookmarks=true]{hyperref}
\usepackage[acronym]{glossaries}
\newacronym{ai}{AI}{Artificial Intelligence}
\newacronym{cfl}{CFL}{Continual Federated Learning}
\newacronym{dnn}{DNN}{Deep Neural Network}
\newacronym{fl}{FL}{Federated Learning}
\newacronym{flfd}{FLfD}{Federated Learning from Demonstration}
\newacronym{hri}{HRI}{Human-Robot Interaction}
\newacronym{iot}{IoT}{Internet of Things}
\newacronym{ml}{ML}{Machine Learning}
\newacronym{ros}{ROS}{Robot Operating System}
\newacronym{svm}{SVM}{Support Vector Machine}

\usepackage{graphicx}
\usepackage{caption}
\usepackage{subcaption}
\usepackage{xcolor}
\begin{document}

\title{Towards Privacy-Aware and Personalised \\ Assistive Robots: A User-Centred Approach}

\author{\authorblockN{Fernando E. Casado}
\authorblockA{Imperial College London\\
fe220@ic.ac.uk}}

\maketitle

\IEEEpeerreviewmaketitle

\section{Motivation}
The demographic landscape is undergoing a significant transformation globally, with the elderly population growing at an unprecedented rate~\cite{grinin2023global}. This shift presents profound challenges in long-term care, as the demand for support for elderly and care-dependent individuals escalates. Traditional caregiving models struggle to meet these demands, leading to an urgent need for innovations that can enhance the quality of life for vulnerable populations while reducing the burden on caregivers.
\emph{Assistive robots} have emerged as a promising solution, offering support in various daily activities. 
Rapid advancements in \acrfull{ml} promise enhanced capabilities for these robots~\cite{gorriz2020artificial,yokoyama2023adaptive,zitkovich2023rt}. By leveraging user and environmental information such as gaze, facial expressions, or nearby objects, \acrshort{ml}-based solutions will enable complex and personalised interactions that foster stronger bonds between users and robots~\cite{mohebbi2020human}. However, as personalised assistive robots approach integration into our daily lives, ethical technology adoption becomes increasingly crucial, with privacy concerns coming to the forefront~\cite{boada2021ethical}.

The widely adopted \acrfull{ros}, as well as related tools, have historically lacked inherent privacy mechanisms~\cite{dieber2017security}. This has placed the responsibility on developers to augment measures through additional hardware and software layers, such as secure communication protocols, user authentication, or sensor obfuscation~\cite{hu2020shadowsense}. 
As robotics increasingly embrace data-driven approaches, the imperative for \emph{privacy by design} is more pressing than ever.
Recent advancements in \acrfull{fl}~\cite{li2021survey} offer a promising avenue for addressing these challenges. 
\textbf{My research aims to pioneer user-centric, privacy-aware technologies like FL, harnessing their potential within assistive robotics to drive impactful advancements and foster societal well-being}.

\begin{figure}[ht]
     \centering
     \begin{subfigure}[b]{0.25\textwidth}
         \centering
         \includegraphics[width=0.99\textwidth]{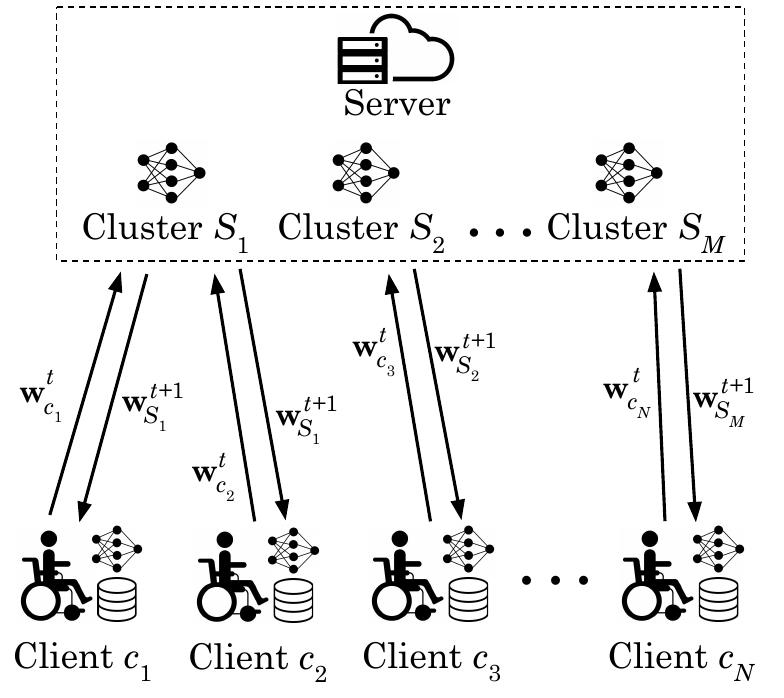}
         \caption{\acrshort{fl} architecture allowing user clustering for personalisation.}
         \label{fig:a}
     \end{subfigure}
     \hfill
     \begin{subfigure}[b]{0.23\textwidth}
         \centering
         \includegraphics[width=0.72\textwidth]{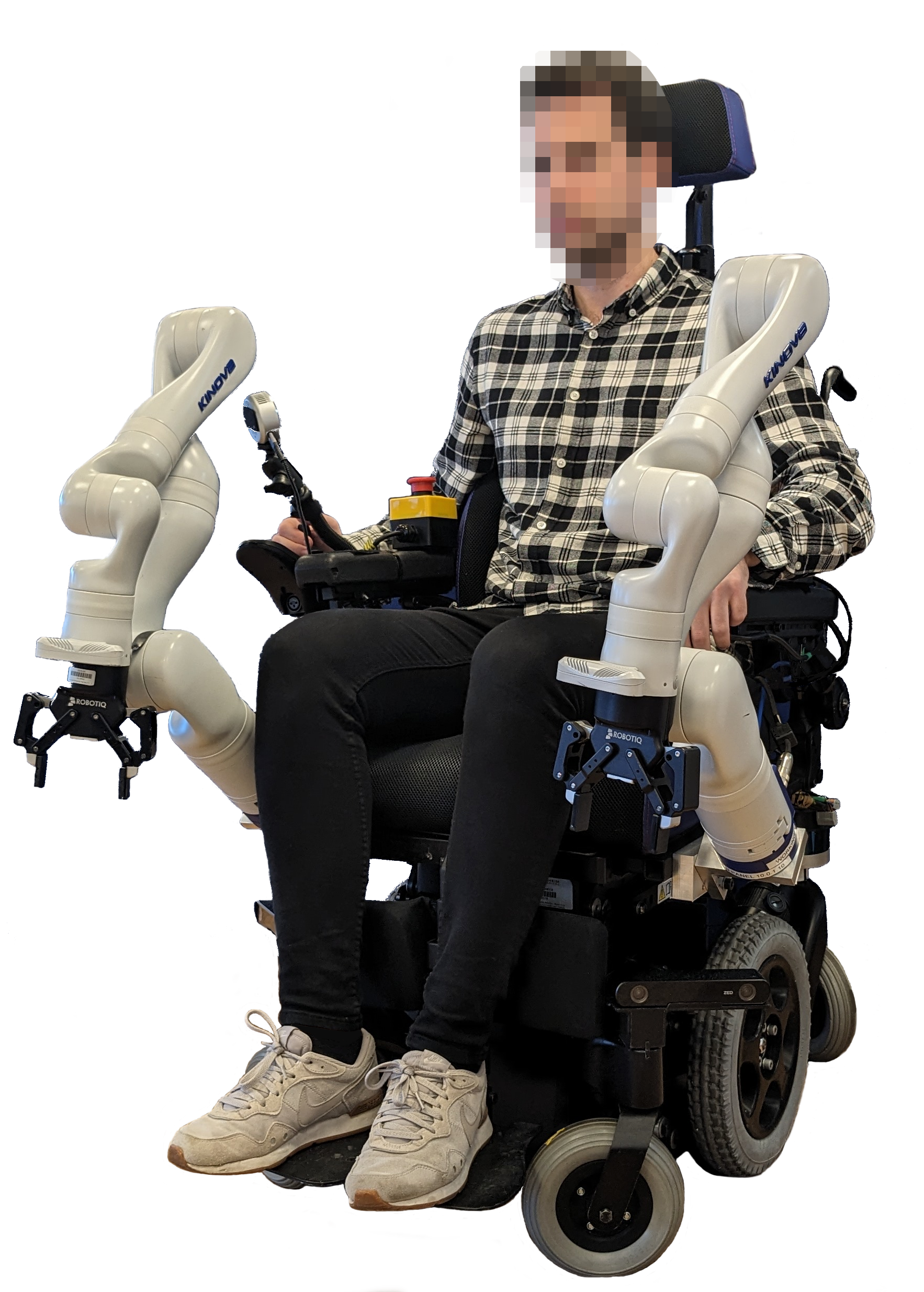}
         \caption{Dual-arm robotic wheelchair for navigation and manipulation.}
         \label{fig:b}
     \end{subfigure}
        \caption{Current work in \acrshort{fl} for wheelchair user assistance.}
        \label{fig:current-work}
    \vspace{-0.5cm}
\end{figure}

\section{Background}

\acrlong{fl}~\cite{criado2021non, li2021survey, mcmahan2017communication} is a young \acrshort{ml} paradigm that enables multiple agents (e.g, robots) to collaboratively learn an inference model from local data without sharing sensitive information with a central server. The learning process is distributed by sharing local parameter updates, ensuring all participants contribute to a robust global model. In this way, \acrshort{fl} mitigates privacy risks and scalability issues associated with centralised data processing. This approach has shown promise in various domains, including medical diagnosis~\cite{brisimi2018federated}, malware detection~\cite{rey2022federated}, or energy demand prediction~\cite{saputra2019energy}.

In the context of assistive robotics, \acrshort{fl} holds significant potential for enabling adaptive and personalised assistance over time. However, deploying \acrshort{fl} solutions in this domain presents unique challenges due to the complex and real-time nature of human-robot interactions and environmental dynamics~\cite{casado2021concept, criado2021non}, along with a still maturing literature that often operates with rather restrictive assumptions. In the following, I detail some of these challenges and outline my previous work in tackling them, aiming to bridge the gap between theoretical advancements and practical implementation.

\subsection{Continual Federated Learning}
A common assumption in \acrshort{fl} is that participants---in our case, human-robot pairs---possess representative and well-labelled datasets from the outset. However, real-world scenarios often differ, with robot sensor data acquired over time. This introduces challenges such as the inability to store and process all information. Additionally, data is often non-stationary, with its distribution evolving unpredictably, a phenomenon known as \emph{concept drift}~\cite{webb2016characterizing}, which may render previously trained models useless.
In our work~\cite{casado2021concept}, we formalised \acrfull{cfl} and proposed a first method to address these challenges. We empirically demonstrated how common \acrshort{fl} approaches can experience performance drops of up to 25\% in non-stationary settings. In contrast, our method integrates mechanisms for detecting and adapting to concept drift, addressing critical questions surrounding the timing and selection of training data. This enables the federated training of \acrfullpl{dnn} from data streams over time without sacrificing performance.
Our work significantly advanced the field of \acrshort{fl}, laying the groundwork for subsequent research tackling similar challenges~\cite{guerdan2023federated, samdanis2023ai, talapula2023hybrid}.

Building upon the prior contribution, in~\cite{casado2023ensemble} we introduced a novel \acrshort{cfl} architecture leveraging ensemble techniques. While \acrshort{fl} usually aligns with \acrshortpl{dnn}, we recognised the importance of algorithmic diversity. Traditional learners, such as Bayesian methods or decision trees, offer distinct advantages, including simplicity and explainability, which are particularly relevant in human-centred robotics. Our architecture embraces non-deep alternatives by posing the model as an ensemble of several independent learners, locally trained. This enables flexible aggregation of diverse local models.

\subsection{Federated Learning for Smart Wheelchair Assistance}
Our work~\cite{casado2022federated} pioneered the application of \acrshort{fl} to solve a real-world problem involving assistive robots, specifically smart wheelchairs. We introduced a comprehensive navigation assistance system to enhance user independence and well-being. Unlike conventional path planning approaches, typically limited by pre-mapped environments, our \acrfull{flfd} framework enables the wheelchair to learn human-like driving skills by direct user involvement. Our decision-making system, where both the user and the model share control, utilises sensitive local data (images and laser readings) to achieve effective navigation after learning from human demonstrations. By incorporating diverse demonstrations from multiple users, our approach addresses the limited generalisation ability in \acrshort{ml}-based literature. Experimental results show superior navigation performance, including tasks like door traversal and obstacle avoidance, outperforming alternative methods in success rate and efficiency.

\subsection{Personalised Federated Learning}
Early \acrshort{fl} proposals advocate generating a global and identical model for all participants. However, the distributed nature of such systems leads to heterogeneous data, posing challenges in effectively combining local contributions. This is especially relevant in assistive robotics, where interactions may occur on different robotic platforms and environments, and where each user has different needs. To address this, the \acrshort{fl} paradigm requires reformulation to personalise the global model while maintaining robustness.
In our work~\cite{criado2021non}, we formally define \emph{heterogeneous data} in \acrshort{fl} and propose a taxonomy covering real-world diversity aspects. Our cohesive framework encompasses two dimensions: spatial (differences among robots, users, and environments) and temporal (\acrshort{cfl}). By analysing requisite conditions for model convergence in various scenarios, we offer novel personalisation strategies, paving the way for individual and group-level adaptations. This foundational work aims to serve as a reference in Continual and Personalised \acrshort{fl}, guiding future research.

\section{Current and future work}
In the trajectory of my research, which initially centred on \acrshort{fl}, there has been a deliberate shift towards addressing tangible challenges in assistive robotics. I aim to advance beyond the current state of the art by developing user-centred technologies that offer both robust and personalised assistance from robots, all while safeguarding privacy.
At present, I am working on new online \acrshort{fl}-based algorithms for dynamic client clustering and personalisation~(see \autoref{fig:a}). The goal is to identify user groups based on custom loss functions utilising personal data collected before and during the interactions. These groups can then engage in cluster-level \acrshort{fl} and even adapt the model to individual needs, striking a balance between generalisation and personalisation.

To demonstrate and validate this research, I am focusing on aiding wheelchair users, given the domain's attention, relevance, and open challenges~\cite{morbidi2023assistive, sivakanthan2022mini}, as well as my prior experience~\cite{casado2022federated}. We are currently developing a new robotic platform, comprising a commercial powered wheelchair equipped with two arms and multiple sensors~(see \autoref{fig:b}). Tasks under exploration include user intention estimation and shared control for concurrent navigation and manipulation, facilitating indoor mobility and interaction. Personalisation efforts aim to enhance the robot's understanding of the user while increasing user autonomy and confidence. This involves considering their physical capabilities, subjective experiences, and preferences.

My interests have also expanded to embrace an interdisciplinary approach, integrating social considerations alongside technical advancements. Recent studies in \acrfull{hri} have brought attention to a privacy-personalisation paradox, with individuals expressing concerns yet willingly sharing information for personalised benefits~\cite{lutz2020robot}. Trust in robots is closely linked to their ability to exhibit privacy-preserving behaviours~\cite{kok2020trust, yang2022effects}, and cultural differences further shape privacy perception, influencing user engagement~\cite{trepte2017cross}. Despite these insights, a gap persists in quantitative evidence and a comprehensive understanding of user privacy perception in \acrshort{hri}.

In future work, I aim to address the questions just mentioned. To this end, I envisage several lines of action. I seek to enhance user awareness of the system's data flow and privacy management, as well as safeguard user-robot communications by mitigating the risk of sensitive information leakage to third parties. To achieve this, I will design novel user interface prototypes utilising diverse peripheral hardware, such as augmented reality headsets. Concurrently, the developed technologies will undergo rigorous assessment through user studies and established questionnaires measuring technology acceptance, cognitive workload, usability, and trust. Recognising the dearth of research in \acrshort{hri} on privacy perception, novel surveys will be devised to evaluate this dimension. A long-term goal is to engage potential end-users, particularly wheelchair users with varying levels of disability, in the evaluation process.

These research objectives are inherently interdisciplinary and ambitious, reaching beyond technological solutions to delve into the complex interactions between humans and assistive robots. In this way, I aim to lead the development of responsible, privacy-aware, user-centred technology, paving the way for the seamless integration of assistive robots into society. Ultimately, my endeavour seeks to enhance the quality of life for elderly and care-dependent individuals.

\bibliographystyle{plainnat}
\bibliography{references}

\end{document}